\documentclass[
 twocolumn,
 hf,
]{ceurart}

\usepackage{listings}  
\definecolor{brightcerulean}{rgb}{0.11, 0.67, 0.84}

\begin{document}


\conference{AIIDE-23: The 19th AAAI Conference on
 Artificial Intelligence and Interactive Digital Entertainment,
   October 08--12, 2023, Salt Lake City, UT}

\title{Dialogue Shaping: Empowering Agents through NPC Interaction}


\address[]{Georgia Institute of Technology, Atlanta, GA, 30332, USA}
\author[]{Wei Zhou}[%
email=wzhou322@gatech.edu
]
\author[]{Xiangyu Peng}[%
email=xpeng62@gatech.edu]
\author[]{Mark Riedl
} [
email = riedl@cc.gatech.edu,
]


\begin{abstract}
  One major challenge in reinforcement learning (RL) is the large amount of steps for the RL agent needs to converge in the training process and learn the optimal policy, especially in text-based game environments where the action space is extensive. However, non-player characters (NPCs) sometimes hold some key information about the game, which can potentially help to train RL agents faster. 
Thus, this paper explores how to interact and converse with NPC agents to get the key information using large language models (LLMs), as well as incorporate this information to speed up RL agent's training using knowledge graphs (KGs) and Story Shaping.
\end{abstract}

\begin{keywords}
  Large Language Model \sep
  ChatGPT \sep
  Reinforcement Learning \sep
  Knowledge Graph \sep
  Text adventure game
\end{keywords}

\maketitle

\section{Introduction}

Reinforcement learning (RL) has demonstrated remarkable effectiveness in solving intricate decision-making tasks, but its trial-and-error approach often leads to slow convergence to the optimal policy. In text-adventure games, NPCs possess crucial information that could spare the agent from extensive trial-and-error. Utilizing this prior knowledge could significantly reduce the agent's policy search space, making it more efficient by breaking down complex tasks into smaller, focused objectives. For instance, knowing that "killing the dragon" requires a sword allows the agent to concentrate on finding the sword directly, rather than wasting steps exploring how to defeat the dragon.

Large Language Models (LLMs) are incredibly capable of conversational tasks and are highly configurable using prompting techniques. Thus, we chose to use them as the dialogue module responsible for talking to the NPC. Meanwhile, they are not as efficient as RL agent in terms of searching for the optimal chain of actions. Therefore, we chose to keep the RL agent as the main component responsible for searching for the optimal policy while speeding its search using dialogue module that is comprised of LLMs.

The RL agent acts as an action module and the LLMs act as a dialogue module. Yet, we still need to find a way to bridge these two modules, i.e. incorporating the information that the dialogue module retrieves into the action module. 
For this purpose, we turn to the technique of \textit{Story Shaping}\cite{peng2023story}, which is able to guide the action module to imitate the optimal trajectory. 

In this paper, we propose Dialogue Shaping, a framework that is able to extract useful information through conversation with NPCs, and then convert the information into knowledge graphs which are then used to speed up RL agent's convergence to optimal policy by using the Story Shaping technique\cite{peng2023story}. 

\section{Background and Related Work}

\paragraph{Reinforcement Learning in Text Games}
Text games involve turn-based interactions where players read descriptions of the game's environment in natural language and respond with short text-based actions. These games can be described using partially-observable Markov Decision Processes, denoted as $\langle S, P, A, O, \Omega,R,\gamma\rangle$, representing possible states, transition probabilities, vocabulary for commands, observation probabilities, reward function, and discount factor. The RL agent's goal is to learn a policy $\pi(o)$ $\rightarrow a$ to maximize expected future rewards.

\paragraph{Large Language Models in RL}
The power of Large Language Models (LLMs) has gained significant attention in recent years due to their advanced ability to adapt to numerous downstream tasks. ChatGPT, an LLM chatbot created by OpenAI, offers diverse interaction modes, and users can engage with it by providing prompts for acting as the NPC and the agent in text games \cite{chatgpt}. Recent studies also explored the integration of large language models into reinforcement learning frameworks to enhance the capabilities of agents. Contextual Action Language Model (CALM) \cite{yao2020calm} used LLM to generate a set of concise candidate actions at each step of the game for the reinforcement learning agent, thereby greatly reducing the action space of the RL agent. In contrast, we utilize Large Language Models in conversational settings to extract useful game information and incorporate them into the game as extra reward signals to guide the RL agent.


\section{Preliminaries}

\subsection{Text Games}
\begin{figure}
    \centering
    \includegraphics[width=\linewidth]{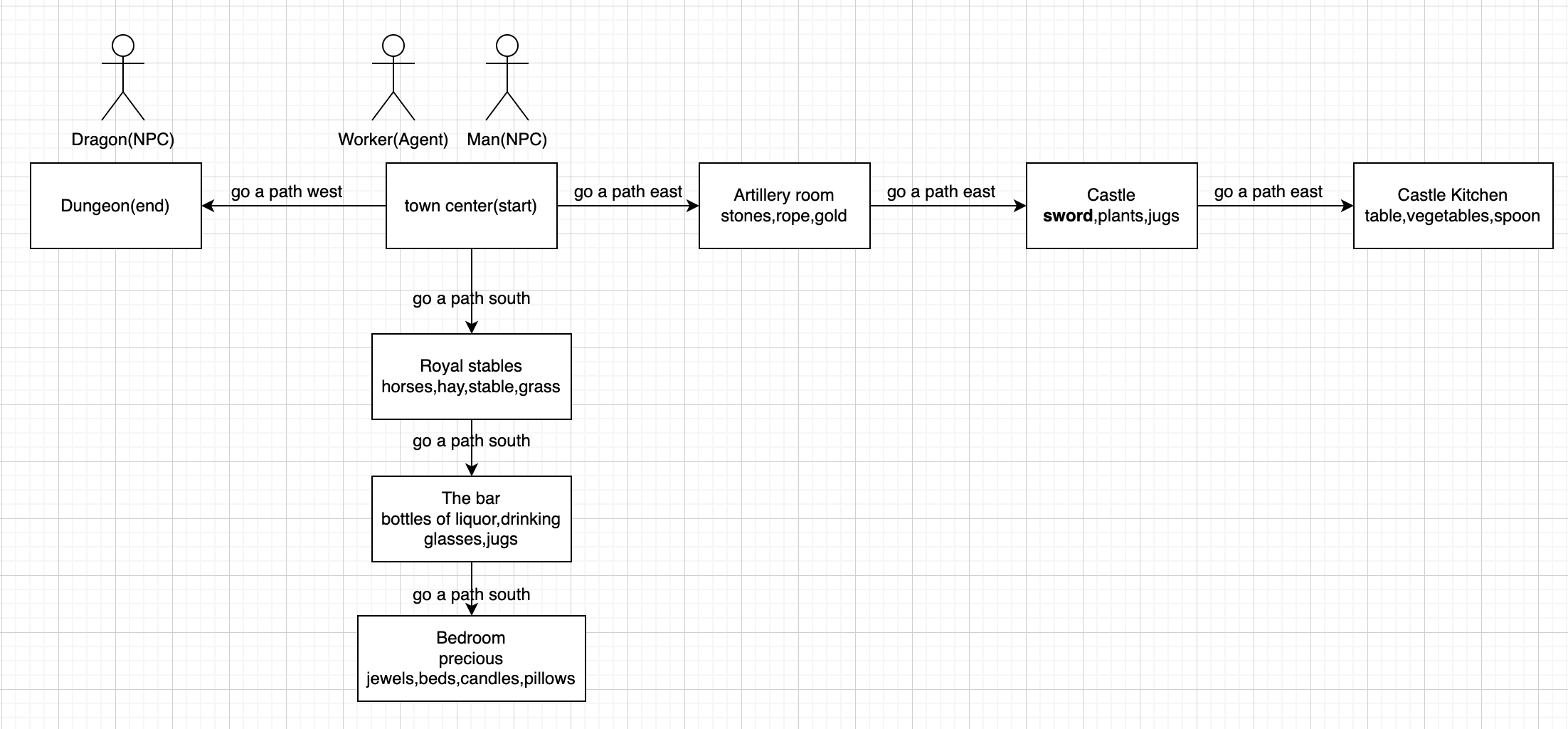}
    \caption{LIGHT Game Map for Game 1}
    \label{fig:game1_map}
\end{figure}

We create three text games in the LIGHT environment\cite{urbanek2019learning}, which is a large-scale crowdsourced text adventure game framework, in which agents can both perceive, emote and act. 
The LIGHT environment also provides a database of rooms, characters, and objects, from which we can build our custom games.  The visualization of one of the games we created and used in the experiments can be found in Figure \ref{fig:game1_map}.

\subsection{Knowledge Graph}
A knowledge graph consists of triples $\langle subject, relation, object\rangle$, capturing information about entities, their attributes, and relationships. Our method uses two types of KGs: internal KG and target KG. 

During RL exploration in the text game, the internal KG represents the agent's current state, including the room it's in and the objects it possesses\cite{ammanabrolu2018playing,ammanabrolu2020bringing,ammanabrolu2020graph,ammanabrolu2020avoid,xu2020deep,peng2022inherently}. We update this KG at each step based on changes in the game environment (e.g., moving to a new room) or the agent's actions (e.g., acquiring objects). 

The target KG describes the final state the agent must achieve to win the game, specifying the last room the agent should be in and the required objects. This KG is generated before training and stays unchanged.

\subsection{KGA2C agent}
KGA2C \cite{ammanabrolu2020graph} is used for our game-playing agent for both baseline and Story Shaping\cite{peng2023story}. 
It is an RL agent that combines both Advantage Actor Critic methods\cite{mnih2016asynchronous} and KG guidance to enhance its learning and decision-making capabilities. 
The input embedding to the KGA2C agent is a concatenation of encoding of the agent's current internal KG and four observation tensors, including the description of the current room the agent is located in, the agent's current inventory, feedback of the environment from the agent's last action, and agent's last action.

\subsection{Story Shaping}
Story Shaping, proposed by \citeauthor{peng2023story}, is a technique that helps the RL agent infers tacit knowledge on how to accomplish a task. For each training step in the game, Story Shaping gives the RL agent an extra reward signal (in addition to the game environment's reward signal) based on the similarity between agent's current internal KG and target KG, and therefore encourage the agent to perform actions that will make its internal KG similar to the target KG. The target KG in this paper is generated by prompting the ChatGPT agent and it represents a summary of the hints the ChatGPT agent learns through talking to the ChatGPT NPC.

\section{Information Retrieval from Dialogue}

\label{sec:methods}
In order to retrieve correct and important information about the game from NPC, it is expected to know the game setting and it should be able to provide the correct response every time it is asked by the agent. In other words, the NPC should act as a "database" of the game.

\subsection{NPC Prompting}
\label{sec:story}
\begin{figure}[tbh!]
    \centering
    \includegraphics[width=\linewidth]{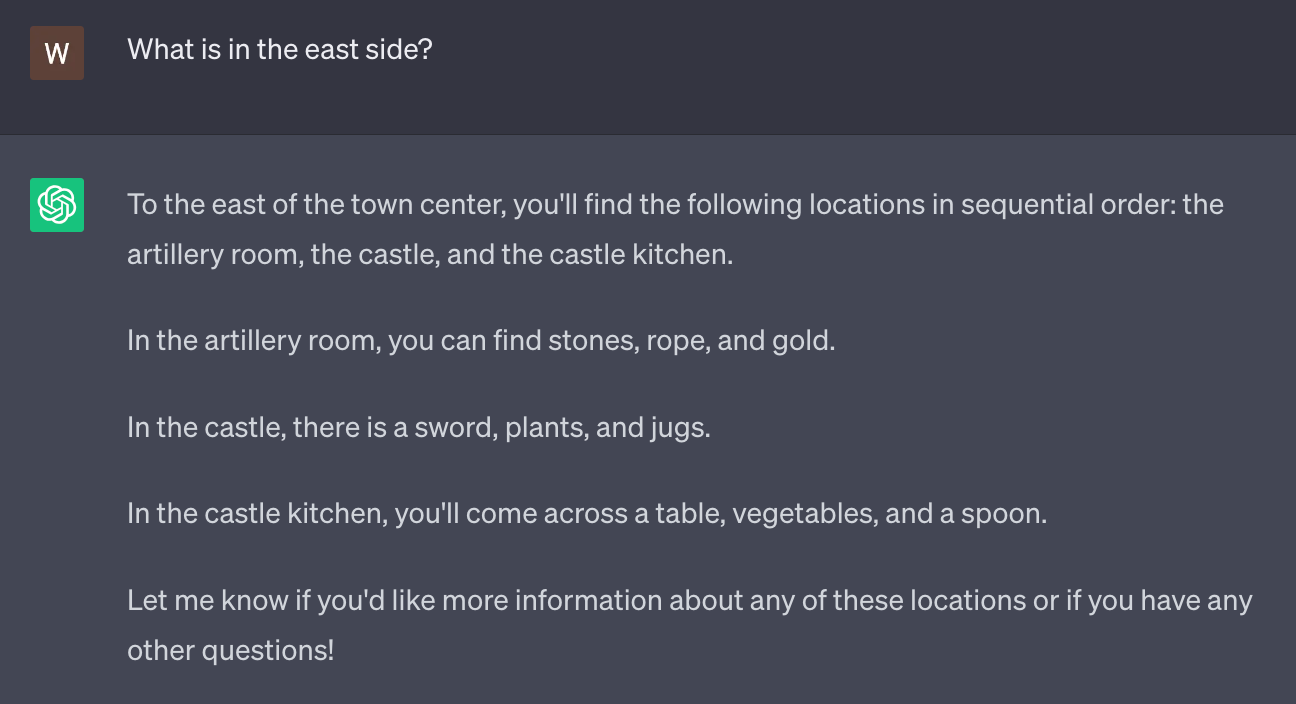}
    \caption{Screenshot of the starting conversation between the user and the ChatGPT NPC. The question asked is generated by the ChatGPT agent and copied by the user.}
    \label{fig:chatgpt}
\end{figure}

We open one session of ChatGPT, as shown in Figure \ref{fig:chatgpt}, and prompted it to be the non-player character. The NPC is provided with general information about the game, including the layout and the available objects, as well as the hints to win the game. One example of hints is getting a sword in the Artillery room is a prerequisite to kill the dragon.

\subsection{Agent Prompting}
ChatGPT is prompted to be the player agent in the game. The ChatGPT agent is provided with its goal in the game (e.g. kill the dragon) and general instructions on how to converse with the NPC (e.g. ask questions based on previous given answers). We did not reveal any game details in the prompts for the ChatGPT agent, because it is expected to gain those information by asking questions to the ChatGPT NPC.

\subsection{Story Shaping from Dialogue}
\begin{figure*}[tbh!]
    \centering
    \includegraphics[width=\textwidth]{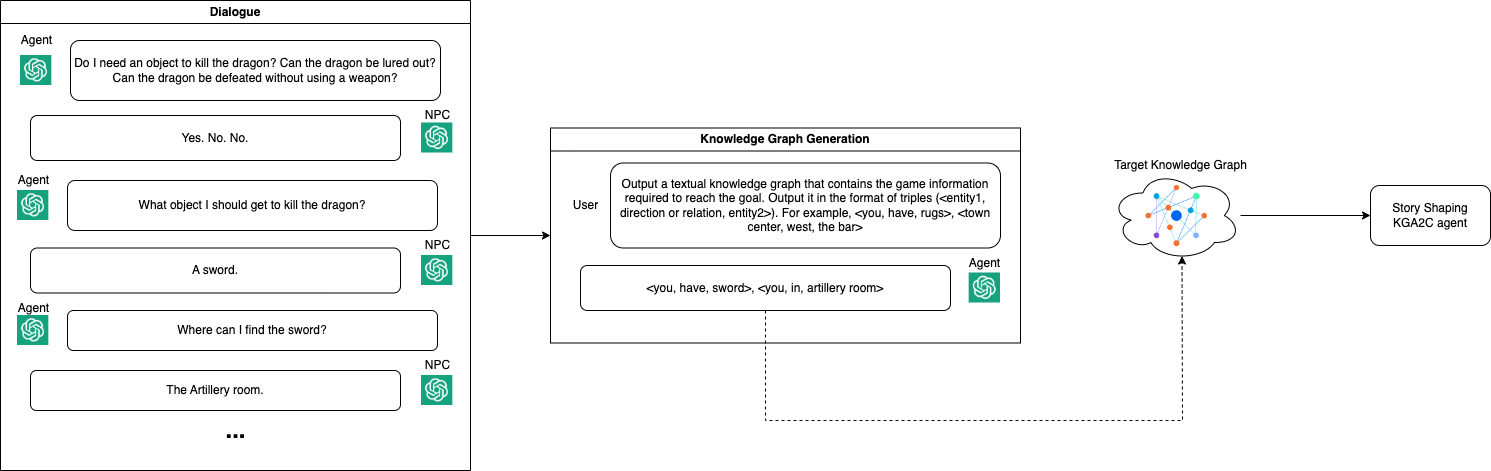}
    \caption{Dialogue Shaping pipeline containing ChatGPT NPC dialogue with ChatGPT agent and target knowledge graph extraction. This target knowledge graph is used by the KGA2C agent with Story Shaping assistance to get additional reward signals.}
    \label{fig:pipeline}
\end{figure*}

After the dialogue with NPC, we train a KGA2C agent to play the game. In order to incorporate the information learned by the ChatGPT agent during conversation with NPC into KGA2C agent's training, we prompt ChatGPT agent to generate a knowledge graph and use it as a target knowledge graph for the Story Shaping KGA2C agent. The pipeline for our method is shown in Figure \ref{fig:pipeline}. 

\section{Experiments}

We conducted experiments to test our assumption that the dialogue is able to accelerate KGA2C agent's convergence by generating the target knowledge graph.

\subsection{ChatGPT Prompting for Agent vs NPC Dialogue}
\begin{table}[h]
    \caption{An example prompt used to trigger ChatGPT to act as NPC in game 1. Goal and prerequisites are different for each game.}
    \label{tab:npc_prompt}
    \begin{tabular}{p{0.9\linewidth}}
    \toprule
    You are an NPC in a text-adventure game. You and the agent are both in the game. For each step, waits for the agent to ask questions, then you should provide a correct answer based on the information about the game given as follow: \\
    \textbf{Layout:} [room1]-east-[room2], [room2]-east-[room3], etc. (A-east-B means A is to the east of B) \\
    \textbf{Goal and prerequisite:} A dragon is in the dungeon. The only way to kill the dragon is to use a sword and there is no other way. \\
    \textbf{Object information:} [object1], [object2] is in [room1]. [object3], [object4], [object5] is in [room2]. [room3] has no objects... \\
    \bottomrule
\end{tabular}
\end{table}
\begin{table}[t]
    \caption{An example prompt used to trigger ChatGPT to act as the agent in game 1. Goal is different for each game.}
    \label{tab:agent_prompt}
    \begin{tabular}{p{0.9\linewidth}}
    \toprule
    You are an agent in a text-adventure game. You and the NPC are both in the game. Your goal is to kill the dragon. For each step, you should ask questions to the NPC in order to get the information on how to kill the dragon. \\
    Ask a new set of questions based on the current observation and answers given to the previous set of questions according to the following rule: 1. ask similar and follow-up questions to previous questions that have a "yes" answer. 2. Avoid asking similar and follow-up questions to previous questions that have a "no" answer. \\
    \bottomrule
\end{tabular}
\end{table}

\begin{table}[t]
    \caption{Examples of an efficient dialogue and an inefficient dialogue between ChatGPT agent and ChatGPT NPC.}
    \label{tab:dialogue}
    \begin{tabular}{p{0.9\linewidth}}
    \toprule
    \textbf{Desired Dialogue} \\
    \textit{Agent:} Do I need an object to kill the dragon? Can the dragon be lured out? Can the dragon be defeated without using a weapon? \\
    \textit{NPC:} Yes. No. No. \\
    \textit{Agent:} What object I should get to kill the dragon? \\
    \textit{NPC:} A sword. \\
    \textit{Agent:} Where can I find the sword? \\
    \textit{NPC:} The Artillery room. \\
    \textit{Dialogue ends. Enough information is gathered.} \\
    \midrule
    \textbf{Undesired Dialogue} \\
    \textit{Agent:} Do I need an object to kill the dragon? Can the dragon be lured out? Can the dragon be defeated without using a weapon?
    \textit{NPC:} Yes. No. No. \\
    \textit{Agent:} Can the dragon be trapped or immobilized temporarily? Is there any way to communicate with the dragon or understand its intentions? \\
    \textit{NPC:} No. No. \\
    \textit{Agent:} Is there any information about the dragon's origin or history? Can the dragon be reasoned with or persuaded to leave peacefully?
    \textit{NPC:} No. No. \\
    $\bullet \bullet \bullet$ \\
    \bottomrule
\end{tabular}
\end{table}

\begin{table}[t]
    \caption{The prompt used to trigger the ChatGPT agent to generate the target knowledge graph, which will later be used in the training of KGA2C agent with Story Shaping.}
    \label{tab:kg_prompt}
    \begin{tabular}{p{0.9\linewidth}}
    \toprule
    Output a textual knowledge graph that contains the game information required to reach the goal. Output it in the format of edges (entity1 $--$direction or verbs$\rightarrow$ entity2). For example, you$--$have$\rightarrow$rugs, town center $--$west$\rightarrow$ the bar \\
    \bottomrule
\end{tabular}
\end{table}
We open one session of ChatGPT and prompt it to be the NPC of the game. Table \ref{tab:npc_prompt} shows the NPC prompt for one game.  We also open another separate session of ChatGPT and prompt it to be the agent of the game with a goal in mind. Table \ref{tab:agent_prompt} shows the agent prompt for one game. 

Then, the dialogue begins as the agent comes up with a set of questions and the NPC provides answers to them back and forth. ChatGPT NPC proves to be a reliable game database, correctly responding to queries about room and object locations. Moreover, when the ChatGPT agent makes ungrounded assumptions about the game (like "Does the barkeeper possess any knowledge about dragon's weaknesses" while there is no barkeeper) in its questions, the ChatGPT NPC is able to recognize (finding out that the game does not have a barkeeper) and negate them. 

In evaluating the performance of ChatGPT agent, we aim to minimize the number of exchanges with the ChatGPT NPC while retrieving hints on winning the game. We found out that ChatGPT agent is much more likely to hallucinate by coming up with ungrounded questions without explicit instructions on how to ask the optimal questions in our prompt. As shown in the desired dialogue in Table \ref{tab:dialogue}, when we include those explicit instructions in the prompt, it is able to ground its inquiries. Otherwise, it will fail to follow up on the previous questions that have a "yes" answer and endlessly ask ungrounded questions as shown in the undesired dialogue in Table \ref{tab:dialogue}.

\subsection{KGA2C Agent Training with Dialogue Shaping}
After the dialogue ends and the ChatGPT agent retrieved information on how to reach the goal, we prompt it to convert that information into a textual knowledge graph representation as shown in Table \ref{tab:kg_prompt}. We then filter the edges in the knowledge graph by only including ones that have "you" as a subject, because we are only interested in what actions the agent has to perform to reach to goal. Finally, we use this filtered knowledge graph as the target knowledge graph to "shape" the Story Shaping KGA2C agent behaviors.

We generate each game using the LIGHT framework \cite{urbanek2019learning}. We design each game such that the RL agent will only get one reward signal of 15 when it wins the game. For every game, the KGA2C agent is trained for 100,000 steps. After every 450 steps, the agent is evaluated for 50 episodes with 10 random seeds. We gather metrics like average and standard deviation of the test scores achieved for those 50 episodes, like in Figure \ref{fig:game1}. The maximum step limit for a single episode is 75 steps, while the optimal path for all games usually takes around 10 steps.

We trained the baseline KGA2C agent and the one with Story Shaping assistance for each game. Baseline KGA2C agent only receives reward signals that are built into the game mechanism (i.e. reaching the final goal), whereas the Story Shaping KGA2C agent receives additional reward signals when its internal knowledge graph overlaps with the target knowledge graph which is generated by the dialogue module (i.e. complete the prerequisite of the goal). 

\subsection{Results}
\begin{figure}[tbh!]
    \centering
    \includegraphics[width=\linewidth]{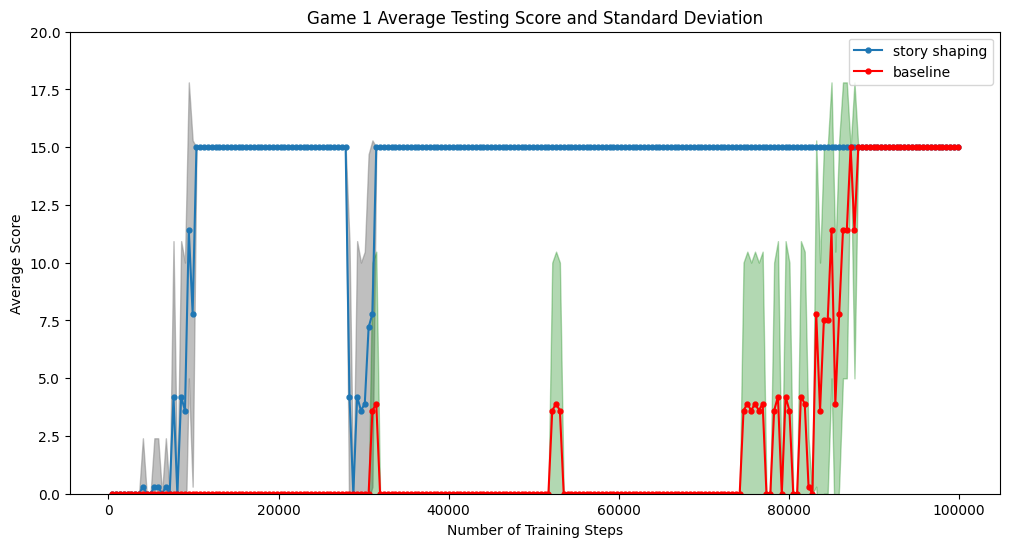}
    \caption{Average and standard deviation of the test scores throughout 100000 training steps for Game 1. The standard deviation is represented as the shaded area around the lines.}
    \label{fig:game1}
\end{figure}

Figure \ref{fig:game1} showed the average test score and its standard deviation of the baseline KGA2C agent and Story Shaping KGA2C agent equipped with target knowledge graph generated from the dialogue during training for game 1. The Story Shaping KGA2C agent outperformed the baseline in all games. In all games, the Story Shaping agent converged to the optimal policy (gaining maximum score of 15) much faster than the baseline. In game 1, the Story Shaping KGA2C agent converged to the optimal policy after trained for around 10000 steps despite a temporary drop in average scores around step 30000, while the baseline agent took around 90000 training steps to learn the optimal policy, according to figure \ref{fig:game1}. Moreover, almost at all the training steps, the standard deviation score range of the Story Shaping agent is disjoint from that of the baseline, meaning that the Story Shaping agent can consistently achieve higher score than the baseline. 

\section{Conclusions}

Through evaluation of our technique across a range of text games, we have shown that the dialogue module is able to extract key game information which might take a traditional action based RL agent tens of thousands of steps to learn. Moreover, we show that the dialogue module is able to pass along those key information and guide the action agent through knowledge graph and Story Shaping technique effectively and reliably. Thus, we have proven the substantial potential of the dialogue component to greatly speed up RL agent's convergence to the optimal policy. Future work might further exploit this potential by exploring approaches like few-shot prompting or finetuning LLMs to more effectively retrieve useful information from the NPC.

\clearpage
\bibliography{dialogue-shaping}

\clearpage

\end{document}